# AI Literacy: An Exercise in Power-Knowledge

**Brady D. Lund[1] and Zoë Abbie Teel[1]**
[1]University of North Texas

**Abstract**

As generative artificial intelligence becomes one of the most significant systems of knowledge production in our society today, questions relating to who can access and shape that production grow increasingly important in our discourse. This paper argues that the existing frameworks for AI literacy, which are dominated by technical competency and responsible-use principles, are insufficient because they enforce a "consumer" orientation toward AI rather than fostering genuine epistemic agency. Based upon Foucault's concept of power-knowledge, Freire's pedagogy of critical consciousness, and scholarship of digital literacy, this paper proposes a reconceptualization of AI literacy as a critical practice that equips individuals not just to use AI systems, but to critically evaluate them, resist their structuring assumptions, and participate in their governance. The paper further argues that unequal access to AI tools in society recapitulates longstanding epistemic injustices, and that a literacy framework oriented toward empowerment must account for these structural inequities. A three-part framework of AI literacy based on the notions of contextual use, critical interrogation, and participatory governance, frame this literacy as a cultivation of epistemic "agents" rather than the training of competent consumer of AI-generated information.

## Introduction

Generative artificial intelligence has emerged not only as a new class of technologies, but also as a new infrastructure for the creation and dissemination of knowledge (Kafedzhiyska 2025). Within a remarkably short period of time, large language models (LLMs) and other generative AI tools have been embedded into our search engines, writing tools, research platforms, and across industries like healthcare and education. These systems have increasingly become used by individuals to access, process, and produce information (Shah 2023). To understand and use generative AI systems has become a foundational competency in modern society.

While the focus on AI literacy has been considerable, the dominant frameworks through which this literacy is currently conceptualized reveal a significant theoretical gap. These frameworks use media literacy, information literacy, and most adjacent digital competency models instrumentally, which reduces complex epistemic and ethical questions to technical proficiency. Most of the existing institutional approaches to AI literacy treat it primarily as a technical skill set: understanding how these models work, evaluating their outputs for accuracy, and using these tools "responsibly" (Lee 2026). This results in an implicit model of the AI-literate person as a competent *consumer*, someone who consumes information from AI systems extensively and efficiently. What is notably absent from these existing AI literacy frameworks is a robust account of power (Stamboliev 2023).

This paper takes the position that the absence of power as a construct in AI literacy discourse is not an incidental omission but rather is symptomatic of a deeper conceptual failure of treating AI systems as neutral tools rather than as what Michel Foucault would recognize as apparatuses of power-knowledge. These are systems that are both productive of knowledge and constitutive of the subjects who encounter that knowledge. When an AI system addresses a medical query or provides an account of a historic incident, it is not just retrieving information as a traditional system might do (Zhao et al. 2023). Rather, it produces a structured account of what is knowable and how. It encodes institutional priorities, disciplinary hierarchies, and the biases of its training data. Those capable of interrogating these functions by understanding not just how to use AI but how AI produces what it presents as knowledge, hold a form of power that those who passively consume AI-generated information do not possess.

As such, AI literacy, properly understood through this lens, is an exercise in power-knowledge. It is not merely a set of technical competencies, nor even a matter of critical thinking about output quality. It is a practice through which individuals develop the capacity to interpret, challenge, and ultimately participate wholly in the governance of algorithmic knowledge production. This capacity is unequally distributed in society, and that inequality is not merely a problem of access but a structural feature of how AI systems are made to reproduce and extend existing epistemic hierarchies.

## Literacy as Empowerment

The term "literacy" is often confused for technical competency. From early debates over print literacy in the 18th and 19th centuries, the question of who should be taught to read – and to what end – has been bound up with questions of social power (Houston 1983).

With Paulo Freire's (1970) *Pedagogy of the Oppressed*, a critical reframing of literacy occurred, with it no longer being framed as the acquisition of technical skills but rather as the development of a critical consciousness. For Freire, the conventional model of literacy education was itself an instrument of oppression: the "banking" model, in which teachers deposit knowledge into passive students, reproduces the very passivity that prevents oppressed people from understanding and challenging the conditions of their oppression. For Freire, genuine literacy requires the ability to name, interrogate, and ultimately transform the social realities encoded in language.

Freirean critical literacy theory has since been extended across a range of domains. In media literacy, scholars such as Len Masterman (2003) argued that functional competency in media consumption – i.e., the ability to read and produce media texts – was insufficient without the critical capacity to interrogate the institutions, economic structures, and ideological assumptions that shape media production. Information literacy frameworks, developed within library and information science, similarly evolved from a skills-based model focused on source evaluation toward more critical frameworks that attend to how information systems themselves structure knowledge access and authority (Tewell 2015). More recently, Delamarter (2025) has extended this critical tradition into the domain of educational technology and algorithmic systems, arguing that frameworks which position learners as passive recipients of algorithmic outputs reproduce what Freire identified with his banking education concept.

## Unique Challenges of AI Literacy

AI systems present distinct challenges for literacy theory. These systems, unlike traditional print or digital media, are not merely channels through which existing knowledge flows to consumers. They are generative systems, producing new knowledge artifacts in response to user inputs (Schwartz & Te'eni 2024). Users of AI can only be considered "literate," then, if they acknowledge their role as co-producer of these knowledge outputs (Velander et al. 2024).

These systems are also opaque in ways that existing information formats are not. The process through which large language models generate responses (i.e., training data, machine learning principles, architectural design), are not accessible to users (Shafik 2026). Even most of the experts designing them are not fully aware of how they work, which is important to consider.

Finally, AI systems are increasingly personalized and dynamic in response to user behavior (Sampson & Bogen 2025). This results in filter effects that shape epistemic habits in ways that are often difficult to detect. These effects extend well beyond the well-known "filter bubble" effect with traditional search engines (Rodilosso 2024). Collectively, these features require a literacy framework that goes beyond critical interpretation of static texts toward an understanding of the relational and process dynamics of AI knowledge production.

## AI Literacy as a Source of Power-Knowledge

At the core of Michel Foucault's (1979) concept of power-knowledge is the idea that knowledge and power are not merely related but constitutively entangled. Systems of knowledge do not simply describe the world but shape it by establishing what can be said, who can say it, and with what authority they can make these decisions. Foucault's (1977) analyses of emergence of new disciplines – medicine, psychiatry, and criminology – show how the emergence of these disciplines occurred simultaneously to the emergence of new forms of power. These forms of power include new methods or technologies for classifying, surveilling, and governing people.

Large language models and generative AI technologies are a new apparatus of power-knowledge within this frame. These models are trained on a vast corpora of human-produced text that encode the full range of institutional, disciplinary, and cultural authorities – the accumulated weight of what has been written, published, indexed, and deemed worth including (Miceli et al. 2022). When these systems generate a response to a user query, they do not merely retrieve information from that corpus; they produce a probabilistically structured account that reflects the distributions of authority and legitimacy within their training data (Gu et al. 2025). The scientific consensus is reflected in the statistical dominance of peer-reviewed text. The cultural mainstream is reflected in the dominance of English-language and Western-produced content (Sukiennik et al. 2025). Institutional authority is reflected in the dominance of formally published sources over marginal, oral knowledge traditions.

This is not a claim that AI systems are simply biased in the pejorative sense, though they are often that as well, but a more fundamental observation: all knowledge systems are selective, and selectivity is always a form of power. The question is not whether AI systems encode power but whether users are equipped to recognize and interrogate that encoding. This incapacity to recognize and name what a system cannot or will not represent is, in Fricker's (2007) terms, a form of *hermeneutical injustice*, one that AI systems risk reproducing and scaling. A user who takes AI outputs as transparent reports on the state of the world is epistemically in a very different position than one who understands those outputs as the products of a particular knowledge apparatus with particular structuring assumptions. The difference between these two users is precisely what a robust AI literacy framework should address.

### Differential Access and the Reproduction of Epistemic Hierarchy

The power-knowledge analysis takes on additional urgency when we examine the differential distribution of AI capabilities. In the Data-Information-Knowledge-Wisdom (DIKW) framework, AI systems do not merely retrieve data but synthesize it into structured knowledge, and increasingly into what might be called *epistemic scaffolding*, the organized frameworks within which individuals understand and reason about complex domains (Peters et al. 2024). Access to this scaffolding is not uniform.

The inequities are multiple and intersecting. Economic inequality structures access at the most basic level: premium AI models (i.e. those with greater reasoning capability, more current knowledge) and more sophisticated domain expertise are available only to paying subscribers (López-Úbeda et al. 2026). Those who cannot afford such subscriptions are confined to less capable free tiers or no access at all, reproducing in the domain of AI what has long characterized access to professional expertise: lawyers, doctors, financial advisors, and researchers have always been differentially available by income. AI does not dissolve this inequality. Instead, it replicates and creates a new layer of it.

Educational capital structures access at a second level. The capacity to interrogate AI systems, in other words, to produce effective prompts, to recognize the limitations and biases of a model's outputs, and to understand when and how to fine-tune or customize models requires a form of meta-cognitive sophistication that is correlated with educational attainment (Draxler et al. 2023). Those with greater formal education are better positioned to exploit AI systems as genuine knowledge tools rather than as simple answer machines. Those with technical training can go further still, accessing model weights, fine-tuning on domain-specific data, and building applications that leverage AI for highly specialized purposes.

Geopolitical inequality structures access at a third level. Consistent high-bandwidth internet connectivity, a prerequisite for effective use of cloud-based AI, is distributed radically unevenly across the globe. This access is disproportionately available in the Global North and in urban centers compared to developing nations and rural areas (Yilmaz 2026). Additionally, the regulatory environments of different nations differentially restrict access to AI tools, and the dominant models are overwhelmingly trained on English-language and Western-produced content, limiting their usefulness for users whose knowledge traditions, languages, and epistemic frameworks are underrepresented in training data.

The cumulative effect of these intersecting inequalities is not just a fairness problem but a structural reproduction of an epistemic hierarchy. Those who already enjoy greater social, economic, and educational capital are better positioned to leverage AI as a tool for the production and deployment of knowledge. Those who lack such capital are more likely to encounter AI as a system that produces knowledge for them, in the passive mode of recipients. This asymmetry precisely recapitulates Foucault's analysis of how knowledge systems produce and reproduce differentially situated subjects (Hannabuss 1996).

### Democratization and Its Contradictions

While the above argument has a clear foundation in the literature, it is important to acknowledge the genuine democratizing potential of AI systems as well. For users with sufficient access and literacy, generative AI does represent a significant redistribution of epistemic resources. Expert knowledge that was previously accessible only through expensive professional services – legal research, medical information, financial analysis, scientific literature synthesis – can now be accessed and interrogated by individuals with modest resources (Dessimoz & Thomas 2024). The ability to generate first drafts, synthesize complex sources, write code, or analyze data without specialized training represents a real expansion of individual epistemic capacity (Lund et al. 2023).

The contradiction, however, is structural. The same features that make AI democratizing for some reproduce hierarchy for others. The productive capacity of AI is a multiplier – it amplifies the existing capabilities of those who can fully access and interrogate it, while offering more modest gains to those who cannot. A researcher with deep domain expertise uses AI to dramatically accelerate their already sophisticated knowledge production; a user with limited education uses AI to get answers to questions, which is useful but does not fundamentally expand their capacity to produce, interrogate, or contest knowledge (Lund et al. 2023). The gap between these users is not closed by AI. To the contrary, it may well be widened. This is the central tension that an adequate AI literacy framework must confront.

## AI Literacy as Empowerment, Not Competency

Current AI literacy frameworks are dominated by what we would describe as a *competency model*: a set of skills and dispositions that enable individuals to use AI tools effectively and responsibly. The competency model is not without value. It is genuinely important that users understand how to evaluate AI outputs, recognize hallucinations, protect their privacy, and avoid over-reliance on AI systems for high-stakes decisions (Lintner 2024). These are real competencies with real social value, especially in a landscape fraught with misinformation and differential AI awareness.

But the competency model is structurally limited by its implicit framing of the AI-literate person as a *consumer*. In competency-based frameworks, the question is always "how do we use AI well?" – where "use" means deriving value from AI outputs and "well" means accurately, responsibly, and in accordance with institutional guidelines. The user is positioned as a consumer of AI-generated knowledge, and literacy is the skill of consuming wisely (Liu et al. 2024). What the competency model does not ask and is not designed to ask is: "how do we understand, interrogate, and shape the systems that produce what AI presents as knowledge?" That omission is not accidental. It reflects the interests of the institutions that produce AI literacy curricula, which are typically the same institutions that create AI systems.

We illustrate this pattern through four publicly available AI literacy frameworks spanning government, higher education, intergovernmental, and global economic institutions: the U.S. Department of Labor's (2026) AI Literacy Framework, Penn State University's (2026) AI Literacy Framework, UNESCO's (2024) AI Competency Framework for Students, and the joint European Commission/OECD AILit Framework, promoted through the World Economic Forum (Milberg 2025). Across all four, the pattern is consistent: competencies are organized around operational proficiency, output evaluation, and responsible use, with no substantive treatment of training data politics, governance participation, or the structural conditions of AI knowledge production. The frameworks differ considerably in their ambition: UNESCO frames students as 'co-creators,' the EC/OECD framework calls for learners to be 'not just consumers of AI, but its ethical stewards,' and Penn State incorporates explicit attention to equity, labor, and democracy. However, all four frameworks converge on an implicit model of the learner as a careful and responsible user of AI outputs rather than a participant in the systems that produce them.

The ideological dimension of this omission can be illuminated by returning to the work of Freire. The competency model of AI literacy is a sophisticated version of the banking model of education: institutions deposit AI skills into users so that users can more efficiently participate in the knowledge economy (Aleman et al. 2025). Users are not invited to examine the conditions under which AI knowledge is produced, the interests it serves, or the alternatives it forecloses. They are trained, as Freire would put it, to adapt to the world rather than to transform it.

### Competency Creates Consumers; Literacy Creates Agents

This paper aims to draw a firm distinction between *consumers* and *agents*. A consumer of AI-generated knowledge is one who receives outputs, evaluates their quality according to established criteria, and acts on them – competently and responsibly, but within a frame of knowledge production that they did not shape and cannot substantially alter. An agent, by contrast, is one who understands the frame itself: who recognizes AI outputs as the products of a particular apparatus of power-knowledge, who can interrogate the assumptions built into that apparatus, who can leverage AI tools not just to receive knowledge but to produce, contest, and share it, and who can participate, to some extent, in the governance of the systems that shape epistemic life.

The agent/consumer distinction is not primarily a distinction of technical sophistication. A highly technically sophisticated user who uses their skills entirely within the logic of AI as a productivity tool is, in the relevant sense, still a consumer. Conversely, a user with modest technical skills who has developed the capacity to interrogate the assumptions of AI outputs – to ask why the system frames a question in a particular way, to recognize whose knowledge traditions are centered and whose are marginal, to seek alternative framings – is exercising a form of epistemic agency that the competency model does not capture (Coeckelbergh 2023).

This distinction has direct implications for how AI literacy should be taught and assessed. A competency-based curriculum will teach students to identify AI-generated misinformation, structure effective prompts, and use AI tools for academic and professional tasks (Long & Magerko 2020). These are valuable skills. An empowerment-based curriculum will additionally teach students to analyze the political economy of AI development, to recognize how training data shapes and limits AI knowledge, to understand the governance mechanisms – or lack thereof – through which AI systems are regulated, and to develop practices of critical co-creation that resist passive consumption of AI outputs (Birkstedt et al. 2023). The empowerment curriculum asks not just "can you use AI well?" but also "do you understand what AI is doing to knowledge, and do you have the capacity to participate in shaping that?"

## A Three-Part Framework for Building AI Literacy That Empowers

In this paper, we propose a framework for AI literacy that is organized around three interlocking dimensions: Contextual Use, Critical Interrogation, and Participatory Governance. These dimensions are not sequential stages but mutually reinforcing concepts that together constitute what we mean by *AI literacy as epistemic agency*. Each dimension corresponds to a distinct relationship between the individual and AI knowledge production: as situated user, as critical interpreter, and as civic participant.

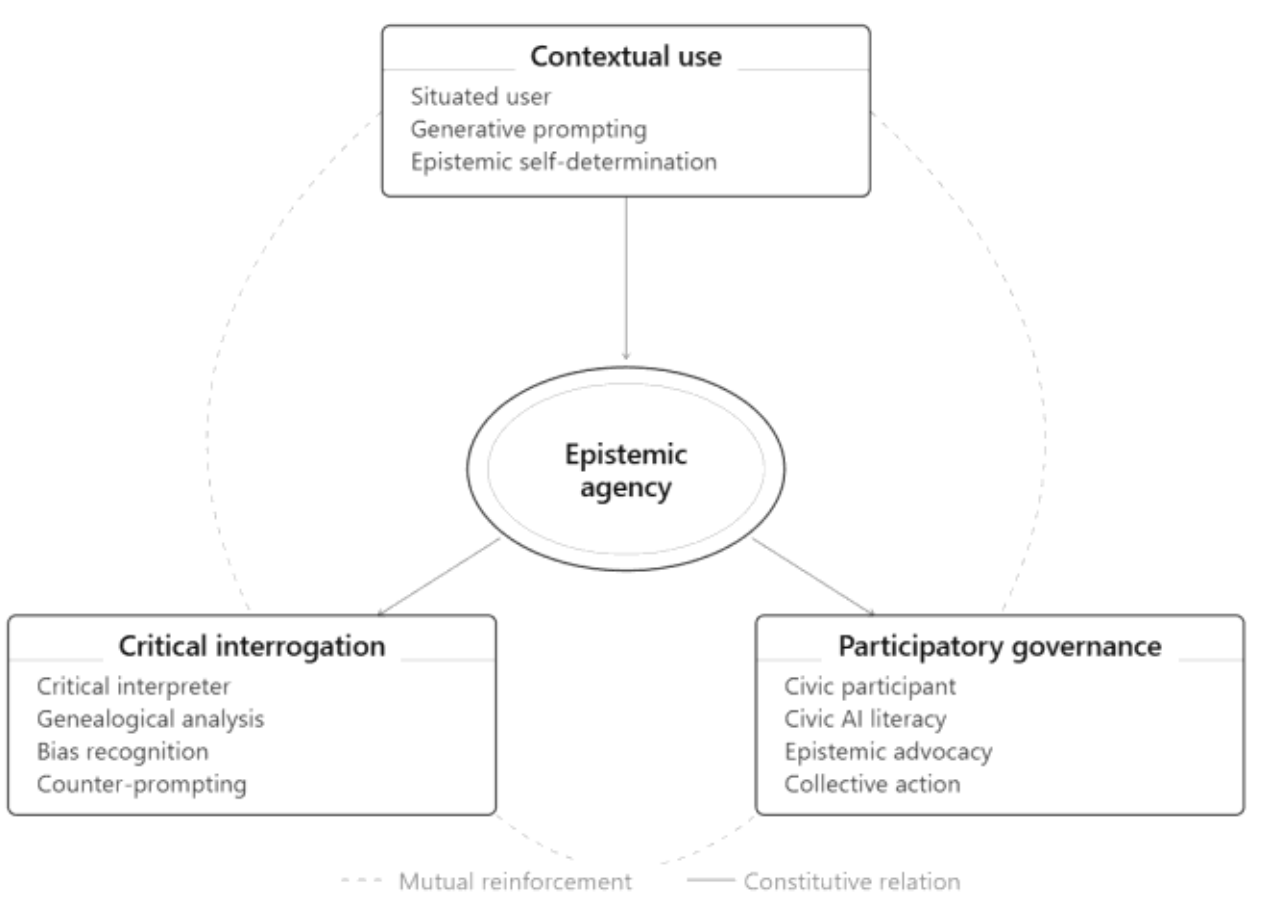

Figure 1: Dimensions of AI Literacy as Epistemic Agency

### Contextual Use

The first dimension, Contextual Use, encompasses what competency-based frameworks capture but reframes it. Contextual Use involves not merely knowing how to operate AI tools but understanding how to use them in ways that serve one's own purposes – where those purposes are understood as extending beyond productivity and efficiency to include epistemic self-determination.

Contextual Use asks: What do I want to know, and why? How does this AI tool construct what it presents as an answer? What are the conditions – e.g., training data, model architecture, deployment context – that shape this output? What is this tool well-suited to and where does it fail? These questions are not purely technical; they require situating AI use within a reflective understanding of one's own epistemic goals and the social contexts in which knowledge matters. A student using AI to research a policy question is engaged in Contextual Use when they understand that the model's framing of the question reflects particular assumptions – about what counts as evidence, whose perspectives are centered – and can account for those assumptions in how they use the output.

Importantly, Contextual Use includes the capacity for what we call generative prompting: using AI not as an answer machine but as a thinking partner, employing iterative dialogue to develop and refine one's own knowledge rather than to outsource it (Helal et al. 2025). This requires understanding AI not as an authority but as a tool with particular capabilities and limitations – a resource for the production of one's own knowledge rather than a substitute for it.

Within the discipline of information science, the concept of contextualized use has existed for decades and has been instrumental in shaping the study of information-seeking behavior. Bates' (1989) seminal berrypicking model established that information use cannot be characterized monolithically. Users do not arrive at information systems with fixed, fully-formed queries that systems then satisfy or fail to satisfy, but with evolving needs and partial understandings that are themselves transformed through the act of seeking. The query is not prior to the search. The query rather emerges throughout the process of searching. Nor can use be value-neutral. To use an AI system is not merely to extract information from it but to enter into a structured relationship with a knowledge apparatus that has already decided – through its training data, its architecture, and its deployment context – what kinds of questions are answerable, what kinds of evidence are authoritative, and what kinds of users it was built to serve. Use, understood in this fuller sense, is always an exertion of power: the power to name what one needs to know, to recognize whether a system's response serves that need or subtly reframes it, and to resist the epistemic closure that comes from accepting a system's framing as the only available one.

### Critical Interrogation

The second dimension, Critical Interrogation, is where the empowerment framework most clearly departs from the competency model. Critical Interrogation involves the capacity to read AI systems as knowledge apparatuses: to understand how they produce the knowledge they produce, whose interests and assumptions are encoded in that production, and what alternatives are foreclosed.

Critical Interrogation encompasses several distinct practices. Genealogical analysis involves tracing the origins of AI outputs – asking not just whether a claim is accurate but how the system came to frame the question in a particular way, what training data and institutional decisions produced this framing, and what different framings were possible (Banerjee 2026). Bias recognition goes beyond identifying factual errors to recognizing the structural patterns (demographic, cultural, disciplinary) through which AI systems reproduce the assumptions of their training corpora (Chuan et al. 2024). Epistemic mapping involves understanding the landscape of what AI systems know and do not know, where their knowledge is deep and where it is shallow, and how to supplement AI outputs with alternative sources, including non-digitized, oral, and vernacular knowledge traditions (Azeroual 2025).

Critical Interrogation also encompasses *counter-prompting*: the practice of deliberately probing AI systems for their assumptions, asking them to represent alternative perspectives, and using their responses as data about the structuring assumptions of the apparatus (Elhosary 2025). A user who asks an AI system to explain the dominant view on a controversial question, then asks it to represent a dissenting view, then asks what perspectives are absent from its training data, is engaged in counter-prompting – using the system reflexively to illuminate its own limitations. This practice transforms the user from a recipient of AI knowledge into an analyst of AI knowledge production.

### Participatory Governance

The third dimension, Participatory Governance, is the most ambitious and the most neglected in existing AI literacy frameworks. It involves understanding AI systems not merely as tools to be used or texts to be interpreted but as sociotechnical systems subject to democratic governance and developing the capacity to participate in that governance.

Participatory Governance encompasses civic literacy about AI: understanding how AI systems are regulated (or not), what institutional interests shape AI development, what mechanisms exist for public accountability, and how individuals and communities can participate in AI governance processes (DiPaola et al. 2024). It involves understanding the political economy of AI – the concentration of AI development in a small number of large corporations, the role of venture capital in shaping research priorities, the relationship between AI companies and state actors – as context for understanding why AI systems are built the way they are (Best et al. 2024).

Participatory Governance also involves *epistemic advocacy*: the capacity to articulate one's interests and values in relation to AI systems and to advocate for AI development that serves those interests. This might range from relatively modest practices – providing feedback to AI systems, participating in public comment processes, supporting organizations working on AI governance – to more ambitious forms of collective action, including labor organizing in AI-affected industries, community-based data governance initiatives, and advocacy for regulatory frameworks that prioritize epistemic justice (Sethy 2026).

Concrete instantiations of participatory governance already exist, though they remain marginal relative to the scale of AI deployment. The EU AI Act's public consultation process, which received over 400 responses from civil society organizations, academic institutions, and affected communities, represents a formal mechanism through which non-expert publics can participate in shaping the regulatory environment for AI systems (Veale & Borgesius 2021). At the community level, initiatives such as the Detroit Digital Justice Coalition's data sovereignty work demonstrate how locally-grounded epistemic advocacy can challenge the assumptions built into algorithmic systems deployed in specific communities (Dev 2019). These examples are imperfect and unevenly resourced, but they illustrate that participatory governance is not merely a theoretical aspiration but a practice that already exists in nascent form and that AI literacy frameworks should equip individuals to join.

The three dimensions of this framework are designed to be cumulative without being in any way hierarchical. Contextual Use is not a precondition for Critical Interrogation, nor Critical Interrogation for Participatory Governance; they are mutually reinforcing. A user who develops critical interrogation skills will be better positioned to engage in contextual use; a user who participates in governance processes will develop a richer understanding of the apparatus they are interrogating. The framework is intended not as a ladder but as a practice, or a set of orientations that together constitute the disposition of the epistemic agent.

## Conclusion

The rapid diffusion of generative AI into the infrastructure of knowledge production is one of the defining developments of the present moment. How society responds to this development – how it conceptualizes the relationship between individuals and AI systems, and what it considers adequate preparation for navigating that relationship – will have lasting consequences for the distribution of epistemic power.

This paper argues that the dominant response, organized around the concept of AI literacy as technical competency, is inadequate for the challenge. The competency model produces what Freire would recognize as adapted subjects – individuals who can navigate AI systems efficiently and responsibly within a frame of knowledge production that they did not shape and cannot substantially contest (Medrado & Verdegem 2024). What is needed instead is an AI literacy framework oriented toward the production of epistemic agents: individuals who understand AI systems as power-knowledge apparatuses, who can interrogate the assumptions encoded in those systems, and who can participate in the democratic governance of AI as a sociotechnical infrastructure.

The three-part framework we have proposed, Contextual Use, Critical Interrogation, and Participatory Governance, is intended as a contribution to this reconceptualization. It is not a complete curriculum or a policy prescription. It is a theoretical frame that we hope can inform the development of more ambitious and more just approaches to AI literacy education. Considerable work remains to be done in translating this framework into a more concrete pedagogical practice, assessing its effectiveness across different educational contexts, and ensuring that it is developed in ways that are responsive to the diverse communities it seeks to serve.

At minimum, it is critical to note the inadequacy of treating AI literacy as a purely technical matter. The question of who can understand, interrogate, and participate in shaping AI systems is a question about the distribution of power in an increasingly AI-mediated epistemic world. It is, in the deepest sense, a political question – and AI literacy discourse that evades that politics is not neutral but complicit. The task of building AI literacy that genuinely empowers requires that we name that politics clearly and build frameworks adequate to it.